\documentclass[a4paper]{article}

\usepackage{INTERSPEECH2022}
\usepackage{kotex}
\usepackage{multirow}
\usepackage{graphicx}
\usepackage{bm}
\usepackage{amsmath}
\usepackage{amssymb}

\title{The Emotion is Not One-hot Encoding:\\ Learning with Grayscale Label for Emotion Recognition in Conversation}
\name{Joosung Lee}
\address{Kakao Enterprise Corp., South Korea}
\email{rung.joo@kakaoenterprise.com}

\begin{document}

\maketitle
\begin{abstract}
In emotion recognition in conversation (ERC), the emotion of the current utterance is predicted by considering the previous context, which can be utilized in many natural language processing tasks. Although multiple emotions can coexist in a given sentence, most previous approaches take the perspective of a classification task to predict only a given label. However, it is expensive and difficult to label the emotion of a sentence with confidence or multi-label. In this paper, we automatically construct a grayscale label considering the correlation between emotions and use it for learning. That is, instead of using a given label as a one-hot encoding, we construct a grayscale label by measuring scores for different emotions. We introduce several methods for constructing grayscale labels and confirm that each method improves the emotion recognition performance. Our method is simple, effective, and universally applicable to previous systems. The experiments show a significant improvement in the performance of baselines.
\end{abstract}

\section{Introduction}
As interest in interactive applications (e.g. chatbots) increases, emotion recognition in conversation becomes more important. Emotions are additional information that can better understand the speaker's state of conversation, and this is used to design an empathic dialogue system~\cite{ERC-research, li-etal-2020-shallow, Lin_Xu_2020}. Emotion also helps provide personalized results, such as social media opinion mining~\cite{chatterjee-etal-2019-semeval} and recommendation systems~\cite{10.1007/978-3-642-38061-7_18}.


In Emotion Recognition in Sentence (ERS) studies, a grayscale label is constructed because one utterance can have multiple emotions. ~\cite{ijcai2018-639} introduces a grayscale label using emotion lexicons such as NRC~\cite{mohammad2013nrc} and Emoticnet~\cite{poria2014emosenticspace}. ~\cite{li2021word} introduces grayscale labels with pre-trained word embeddings. However, since these methods do not consider utterance, there is a limitation that one-hot encoding is mapped to one grayscale label. ~\cite{Guo_Han_Han_Huang_Lu_2021} trains the model with grayscale labels through the basic predictor. However, it depends on the basic predictor with poor performance and cannot handle the noise.

In emotion recognition in conversation (ERC), an utterance has a more sensitive distribution of emotions than in ERS because emotions must be recognized in consideration of both context and utterance. However, most ERC datasets are labeled for only one emotion, and the latest models are trained with labels of one-hot encoding. That is, when the utterance \textit{"yeah, i do"} contains both \textit{joy} and \textit{neutral} emotions depending on the context, learning with only one emotion can cause an error. Our paper focuses on solving the limitation that a model is trained with only one emotion, but it is very expensive for humans to relabel multiple emotions in every utterance. Therefore, we construct grayscale labels in several automatic ways. We propose the following methods for constructing grayscale labels: 1) \textbf{Category} 2) \textbf{Word-Embedding} 3) \textbf{Self} 4) \textbf{Self-Adjust} 5) \textbf{Future-Self}. "Cateogry" and "Word-embedding" are mapping one-hot encodings to grayscale label (i.e. soft-label encoding), which means that training samples with the same emotion are always mapped to the same grayscale label. "Self"-type methods utilize a pre-trained self-teacher-model to construct grayscale according to utterances, which are mapped to different grayscale labels regardless of ground-truth emotion. To the best of our knowledge, our work is the first attempt to create multiple grayscale labels in ERC.

We construct grayscale labels to four ERC datasets: IEMOCAP, dailydialog, MELD, and EmoryNLP. Experimental results show that RoBERTa-large~\cite{RoBERTa} achieves competitive performance with only grayscale labels, which is simple but effective. We also show that grayscale labels can be used in the latest approaches, Psychological~\cite{li-etal-2021-past-present}, CoG-BART~\cite{li2021contrast}, DAG-ERC~\cite{shen-etal-2021-directed} and CoMPM~\cite{lee2021compm}.

\begin{figure*}[!t]
    \centering 
    \includegraphics[width=1.4\columnwidth]{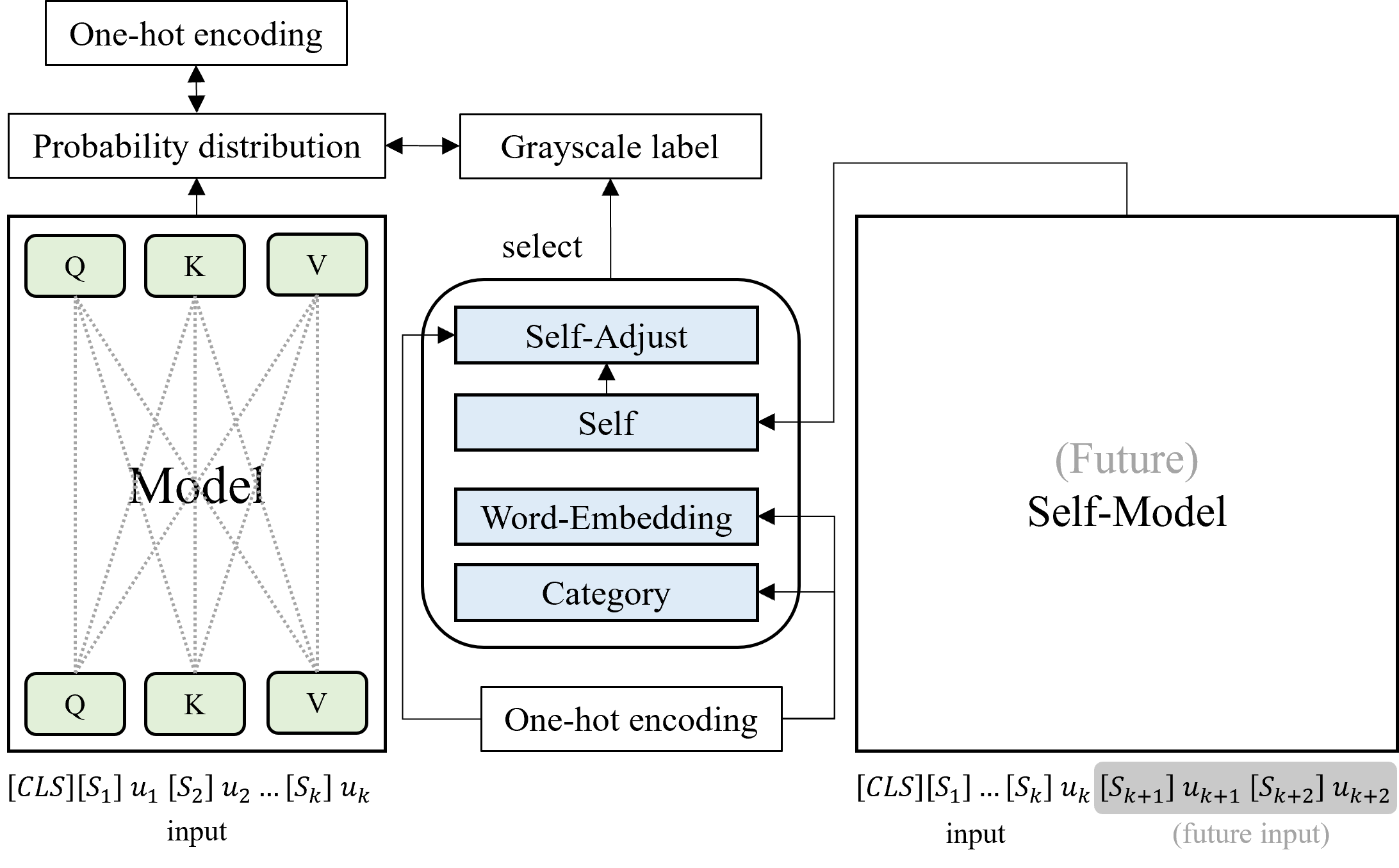}
    \caption{An overview of our framework based on RoBERTa. For each input, a grayscale label is constructed, and the model is trained so that the distance between the probability distribution of emotion and the distribution of the grayscale label is close.}
    \label{fig:Model}
\end{figure*}

\section{Proposed Approach}
\subsection{Overview}
Figure~\ref{fig:Model} is an overview of our approach to recognizing the emotion of the $k$-th utterance $u_k$. First, we construct grayscale labels on our training samples in various ways. Then, the model is trained on both the one-hot encoding and the grayscale label. The input is concatenated with the previous context and the current utterance, and special tokens indicate the speaker by prepending each utterance. Also, we prepend the $[CLS]$ token to the input in RoBERTa. The model is trained to predict emotions with the $[CLS]$ token.

\subsection{Construction of Grayscale Label}
\label{sec:grayscale}
We propose 5 ways to automatically construct grayscale labels. The "category" method constructs a grayscale label heuristically. The "word-embedding" method constructs a grayscale label using word-embedding. Other methods (i.e. self-methods) construct grayscale using logit of the self-model (i.e. teacher-model).

\subsubsection{Category}
\cite{ijcai2018-639} generates grayscale labels as a strategy based on linguistic resources, which uses words from each sentence and ground-truth together. Inspired by this, using only the ground-truth, we simply divide emotions into sentiment categories and score emotions differently for each category. Categories are shown in Table~\ref{tab:dataset_cate}. The scores ($s_i$) of emotions ($e_i$) are as follows:



\begin{equation}
  s_i =
    \begin{cases}
      1 & \text{if } e_i = e_{gt}\\
      0.5 & \text{if } \text{category}(e_i) = \text{category}(e_{gt})\\
      0 & \text{otherwise}
    \end{cases}       
\label{eq:category_score}    
\end{equation}
where $e_{gt}$ is the the ground-truth emotion. In the category method, emotions included in the same sentiment give positive scores to each other. We calculate the grayscale label $\mathbf{g} = \{g_1, g_2, ..., g_k \}$ ($k$ = the number of class) by normalizing $\mathbf{s}$ as follows:
\begin{equation}
\label{eq:grayscale_noramlizing}
g_i = \frac {s_i} {\sum_{j=1}^{k} s_j}
\end{equation}

\subsubsection{Word-Embedding}
Inspired by ~\cite{li2021word}, we calculate the similarity between emotions using FastText~\cite{mikolov2018advances}, a publicly released word-embedding~\footnote{We used FastText, but other public word-embeddings are fine. A more advanced method is to use word-embedding learned from data such as emotion lexicon.}. Word-embedding refers to the latent representation of a word. Therefore, the similarity between words with similar meanings is high, and the similarity between words with different meanings is low. The score for each emotion is calculated using cosine similarity as follows:

\begin{equation}
s_i = \mathbf{max} (\frac {w_i \cdot w_{gt}} {\left \| w_i \right \| \left \| w_{gt} \right \|}, 0)
\label{eq:word_score}    
\end{equation}
where $w_{gt}$ is the embedding vector of the ground-truth emotion. 

We construct the grayscale label $\mathbf{g}$ in two steps. First, if the similarity score is negative, it is changed to 0. Because a negative cosine similarity value indicates a less relevant emotion, the $s_i$ of the unrelated emotion is changed to 0 before linear normalizing. Then, the grayscale label is calculated through normalizing as in Equation~\ref{eq:grayscale_noramlizing}. The spelling for each emotion is explained in Table~\ref{tab:dataset_cate}.


\subsubsection{Self}
The "category" and "word-embedding" methods have a limitation in that grayscale labels are calculated only in the relation to emotions without considering utterances. Also, the word-embedding method has a disadvantage in that the embedding can be similar even if the meaning of the word is different (e.g \textit{night} and \textit{day}). So we propose the self-method, which is similar to~\cite{Guo_Han_Han_Huang_Lu_2021} in ERS. The self-grayscale label is constructed from the emotion of the utterance predicted by the self-model (fine-tuned model), which is a kind of distillation~\cite{Gou2021}. Since distillation is not our goal, we use a self-model instead of exploring a new teacher-model. 

First, the teacher-model is trained as a classification task using one-hot encoding. The logit of the pre-trained teacher-model becomes the score, and then the grayscale label is calculated through the softmax as follows:

\begin{equation}
g_i = \frac {e^{s_i}} {\sum^{k}_{j=1} e^{s_j} }
\label{eq:self_grayscale}    
\end{equation}

When the student-model is trained, the parameters of the teacher-model are fixed, where the student-model is the final model trained with ERC data (RoBERTa). Each training sample has a different grayscale label according to the utterance regardless of the ground-truth emotion.

\subsubsection{Self-Adjust}
\begin{figure}[!t]
    \centering 
    \includegraphics[width=1.0\columnwidth]{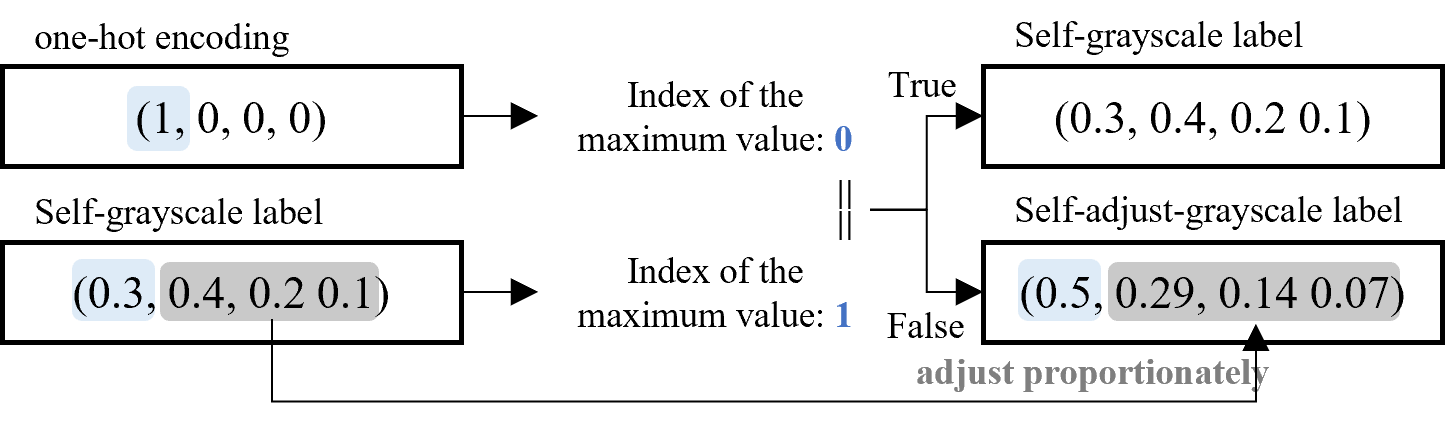}
    \caption{The example of constructing a self-adjust-grayscale label.}
    \label{fig:adjust}
\end{figure}

The self-method depends on the performance of the self-model. So, for example, if the one-hot encoding of the ground-truth is (1, 0, 0, 0) and the self-grayscale label is (0.3, 0.4, 0.2, 0.1), there is a risk that the model is confused about the best emotion. To alleviate this problem, we propose a self-adjust-method to adjust the self-grayscale label. Figure~\ref{fig:adjust} is an example of the self-adjust-method.

If ground-truth emotion is different from the maximum probability of the self-grayscale label, we adjust as follows: 1) The self-adjust-grayscale value corresponding to the gold emotion is adjusted to 0.5. 2) Other emotions have a value divided by 0.5 as much as the distribution of self-grayscale labels. The adjusting function is calculated as follows:

\begin{equation}
  g'_i =
    \begin{cases}
      0.5 & \text{if } i = gt \\
      \frac {0.5 g_i} {1 - g_{gt}} & \text{otherwise}
    \end{cases}       
\label{eq:sa_grayscale}    
\end{equation}
where $gt$ is the index corresponding to the ground-truth emotion, and $g'_i$ is the adjusted grayscale label.

\subsubsection{Future-Self}
We improve the performance with an additional strategy when a self-model is trained. In ERC research, it is common to predict the emotion of an utterance by considering only the past context. If the model captures information in a future context, it cannot predict emotions in real-time during inference. A non-real-time emotion recognition system has the disadvantage of being difficult to be utilized in other natural language processing tasks. However, it can be helpful in the training phase because the future context has information about the current emotion that affects the listener's response. 

Therefore, we propose a method to improve the performance of the self-model by using the future context as input. The input is concatenated with the utterances of two future turns, and this model is called the future-self-model. 

\subsection{Loss}
We train the model on both a one-hot encoding and a grayscale label. The one-hot encoding is used to predict the top one of emotions, and the grayscale label is used to predict the probability distribution reflecting the distance between emotions. Both tasks use cross-entropy loss:

\begin{align}
    L_b &= -\frac{1}{N} \sum^{N}_{n=1} \sum^{k}_{i=1} o_i log(p_i) \\
    L_g &= -\frac{1}{N} \sum^{N}_{n=1} \sum^{k}_{i=1} g_i log(p_i) \\
    L &= L_b + \alpha L_g
\label{eq:loss}
\end{align}
where $N$ is the number of training data, $k$ is the number of emotion classes, $o_i$ is a one-hot encoding, $g_i$ is a grayscale label, and $p_i$ is the emotion probability predicted by the model. $\alpha$ is described in Section~\ref{sec:alpha}.

\begin{table}[!t]
\centering
\resizebox{1.0\columnwidth}{!}{
\begin{tabular}{c|ccc|ccc}
\hline
\multirow{2}{*}{Dataset} & \multicolumn{3}{c|}{dialogues}                                & \multicolumn{3}{c}{utterance}                                \\ \cline{2-7} 
                         & \multicolumn{1}{c|}{train} & \multicolumn{1}{c|}{dev}  & test & \multicolumn{1}{c|}{train} & \multicolumn{1}{c|}{dev}  & test \\ \hline
IEMOCAP                  & \multicolumn{1}{c|}{108}   & \multicolumn{1}{c|}{12}   & 31   & \multicolumn{1}{c|}{5163}  & \multicolumn{1}{c|}{647}  & 1623 \\ \hline
DailyDialog              & \multicolumn{1}{c|}{11118} & \multicolumn{1}{c|}{1000} & 1000 & \multicolumn{1}{c|}{87170} & \multicolumn{1}{c|}{8069} & 7740 \\ \hline
MELD                     & \multicolumn{1}{c|}{1038}  & \multicolumn{1}{c|}{114}  & 280  & \multicolumn{1}{c|}{9989}  & \multicolumn{1}{c|}{1109} & 2610 \\ \hline
EmoryNLP                 & \multicolumn{1}{c|}{713}   & \multicolumn{1}{c|}{99}   & 85   & \multicolumn{1}{c|}{9934}  & \multicolumn{1}{c|}{1344} & 1328 \\ \hline
\end{tabular}
}
\caption{Statistics for 4 datasets.}
\label{Tab:dataset}
\end{table}

\section{Experiments}
In the "category" and "word-embedding" methods, the model is jointly trained with a one-hot encoding and a grayscale label as Equation~\ref{eq:loss}. In the self-methods, first, the teacher-model is trained as a classification task using one-hot encoding. Then, the parameters of the teacher-model are frozen and the student-model is jointly trained using one-hot encoding and grayscale labels as Equation~\ref{eq:loss}. 

\subsection{Training Setup}
\label{sec:setup_appendix}
We use the pre-trained model from the huggingface library~\footnote{https://github.com/huggingface/transformers}. The optimizer is AdamW and the learning rate is 1e-6 as an initial value. The learning rate scheduler used for training is \textit{get\_linear\_schedule\_with\_warmup}, and the maximum value of 10 is used for the gradient clipping. We select the model with the best performance on the validation set. All experiments are conducted on one A100 GPU.

\subsection{Dataset and Evaluation}
We experiment on four datasets: IEMOCAP~\cite{iemocap}, DailyDialog~\cite{dailydialog}, MELD~\cite{poria-etal-2019-meld}, and EmoryNLP~\cite{emorynlp}. Table~\ref{Tab:dataset} shows the statistics of the data. DailyDialog uses 7 classes for training, but we measure Macro-F1 for only 6 classes excluding neutral. Other datasets are evaluated with weighted average F1. 

\begin{table}[!t]
\centering
\resizebox{1.0\columnwidth}{!}{
\begin{tabular}{c|ccc}
\hline
\multirow{2}{*}{Dataset} & \multicolumn{3}{c}{category}                                                                                         \\ \cline{2-4} 
                         & \multicolumn{1}{c|}{positive}                  & \multicolumn{1}{c|}{negative}                    & neutral           \\ \hline\hline
IMEOCAP                  & \multicolumn{1}{c|}{excited, happy}            & \multicolumn{1}{c|}{angry, frustrated,   sad}    & neutral           \\ \hline
DailyDialog              & \multicolumn{1}{c|}{happy}                     & \multicolumn{1}{c|}{anger, disgust, fear,   sad} & neutral, surprise \\ \hline
MELD                     & \multicolumn{1}{c|}{joy}                       & \multicolumn{1}{c|}{anger, disgust, fear,   sad} & neutral, surprise \\ \hline
EmoryNLP                 & \multicolumn{1}{c|}{joy, peaceful,   powerful} & \multicolumn{1}{c|}{mad, sad, scared}            & neutral           \\ \hline
\end{tabular}
}
\caption{The sentiment category of the emotion corresponding to each dataset}
\label{tab:dataset_cate}
\end{table}

In IMEOCAP, the emotional inventory (6) is given as "happy, sad, angry, excited, frustrated and neutral". In DailyDialog, the emotional inventory (7) is given as "anger, disgust, fear, joy, surprise, sadness and neutral". In MELD, the emotional inventory (7) is given as "anger, disgust, sadness, joy, surprise, fear and neutrality". In EmoryNLP, the emotional inventory (7) is given as "joyful, peaceful, powerful, scared, mad, sad and neutral". Table~\ref{tab:dataset_cate} shows the sentiment category of each emotion, and the corresponding word-embedding is used.

\begin{table*}[!htb]
\centering
\resizebox{1.5\columnwidth}{!}{
\begin{tabular}{c|c|cc|c|c}
\hline
\multirow{2}{*}{Models}   & IEMOCAP        & \multicolumn{2}{c|}{DailyDialog} & MELD             & EmoryNLP         \\ \cline{2-6} 
                          & W-Avg F1       & Macro F1        & Micro F1       & W-Avg F1 & W-Avg F1 \\ \hline\hline
RoBERTa (ours)                   & 63.17          & 51.17           & 58.63          & 64.79            & 36.3            \\
+C   (category)          & 64.83          & 53.44           & 58.91          & 65.27            & 36.97            \\
+W   (word-embedding)               & 64.85          & 51.77           & 60.15          & 65.48            & 36.05            \\
+S   (self)               & 64.78           & 53.89           & 60.67          & 65.71            & 29.09            \\
+SA   (self-adjust)         & \textbf{65.85}          & 53.75           & 59.73          & 66               & 37.69            \\
+FSA   (future-self-adjust) & -              & \textbf{55.84}  & \textbf{61.67} & \textbf{66.49}            & \textbf{38}               \\
\hline\hline
Psychological~\cite{li-etal-2021-past-present}                   & 63.37 & 52.38               & 60.54          & 64.38            & 37.32            \\

+S                        & \textbf{65.43} & 53.63     & 59.98  & \textbf{65.24}   & \textbf{38.96}            \\
+SA                       & 64.92          & \textbf{55}          & \textbf{60.79}          & 65.04            & 38.33            \\
\hline\hline
DAG-ERC~\cite{shen-etal-2021-directed}                   & 68.03 & -               & 59.25          & \textbf{63.63}            & 39.02            \\

+S                        & \textbf{68.57} & -     & \textbf{59.4}  & 63.55   & \textbf{40.23}            \\
+SA                       & 68.43          & -          & 59.33          & 63.57            & 39.85            \\
\hline\hline
CoG-BART~\cite{li2021contrast}                   & 63.8 & 53.99               & 55.52          & 64.77            & 36.48            \\

+S                        & 64.55 & \textbf{57.14}     & \textbf{57.15}  & 64.26   & 36.97            \\
+SA                       & \textbf{64.77}          & 56.07          & 56.57          & \textbf{65.28}            & \textbf{37.42}           \\
\hline\hline
CoMPM~\cite{lee2021compm}                   & 66.33 & 52.46               & 60.41          & 65.53            & 38.56            \\

+S                        & 67.5 & \textbf{52.88}     & \textbf{61.1}  & 66.19   & 39.05            \\
+SA                       & \textbf{67.52}          & 52.59          & 60.66          & \textbf{66.5}            & \textbf{39.95}            \\
\hline
\end{tabular}
}
\caption{Our results are the average of three runs. The performance of the comparison system is the result of our re-experiment with the published code. Bold text indicates the best performance in each part.}
\label{tab:results}
\end{table*}

\subsection{Results and Discussion}
Table~\ref{tab:results} shows our experimental results. First, our proposed RoBERTa has slightly improved performance by using special tokens compared to the RoBERTa proposed in previous studies. RoBERTa trained with grayscale labels generally improves performance. However, the performance doesn't improve according to the combination of data and grayscale type (i.e. EmoryNLP+(W or S)). In particular, +S negatively affects RoBERTa in EmoryNLP. We infer that +S contains a lot of noise due to the insufficient performance of the self-model in EmoryNLP. +SA proposed to alleviate the noise problem is effective in improving the performance. In RoBERTa, we observe that self-methods are generally superior to +C and +W. +C and +W are very simple and effective, but there is a limit that does not consider utterance. However, the self-methods considering dialogue further improve performance because they are more fine-grained grayscale labels. Because future-self-RoBERTa is inferior to original RoBERTa in IEMOCAP, RoBERTa+FSA was not tested. In other datasets, +FSA outperforms +SA because the future-self-model constructs fine-grained grayscale labels more than the self-model.

We apply +S and +SA, which were effective in RoBERTa, to other models. In other models, unlike RoBERTa, +FSA is not used because future inputs cannot simply be combined with the models. +S and +SA give a positive signal to the performance of the other models. In addition, unlike RoBERTa, +S also improves model performance on EmoryNLP. Analysis of this should be explored, but we assume that (teacher-) comparative systems learned about EmeryNLP do not have false biases, unlike RoBERTa. Therefore, the proposed grayscale labels can improve the average performance regardless of the model structure.

Instead of focusing on modeling, our research focuses on constructing a grayscale label with the distribution of emotions contained in the utterance. We achieve high performance competitive with the state-of-the-art models just by combining grayscale with RoBERTa. We also show performance improvements by combining grayscale with the original state-of-the-art models. In other words, grayscale labels are easy to effectively combine with other models. However, since grayscale labels have different effects depending on the combination of model and data, it is difficult to select the optimal grayscale label and $\alpha$ for the model through several experiments. Exploring $\alpha$ is shown in the next section.


\subsection{Performance According to $\alpha$}
\label{sec:alpha}

\begin{figure}[!t]
    \centering 
    \includegraphics[width=0.85\columnwidth]{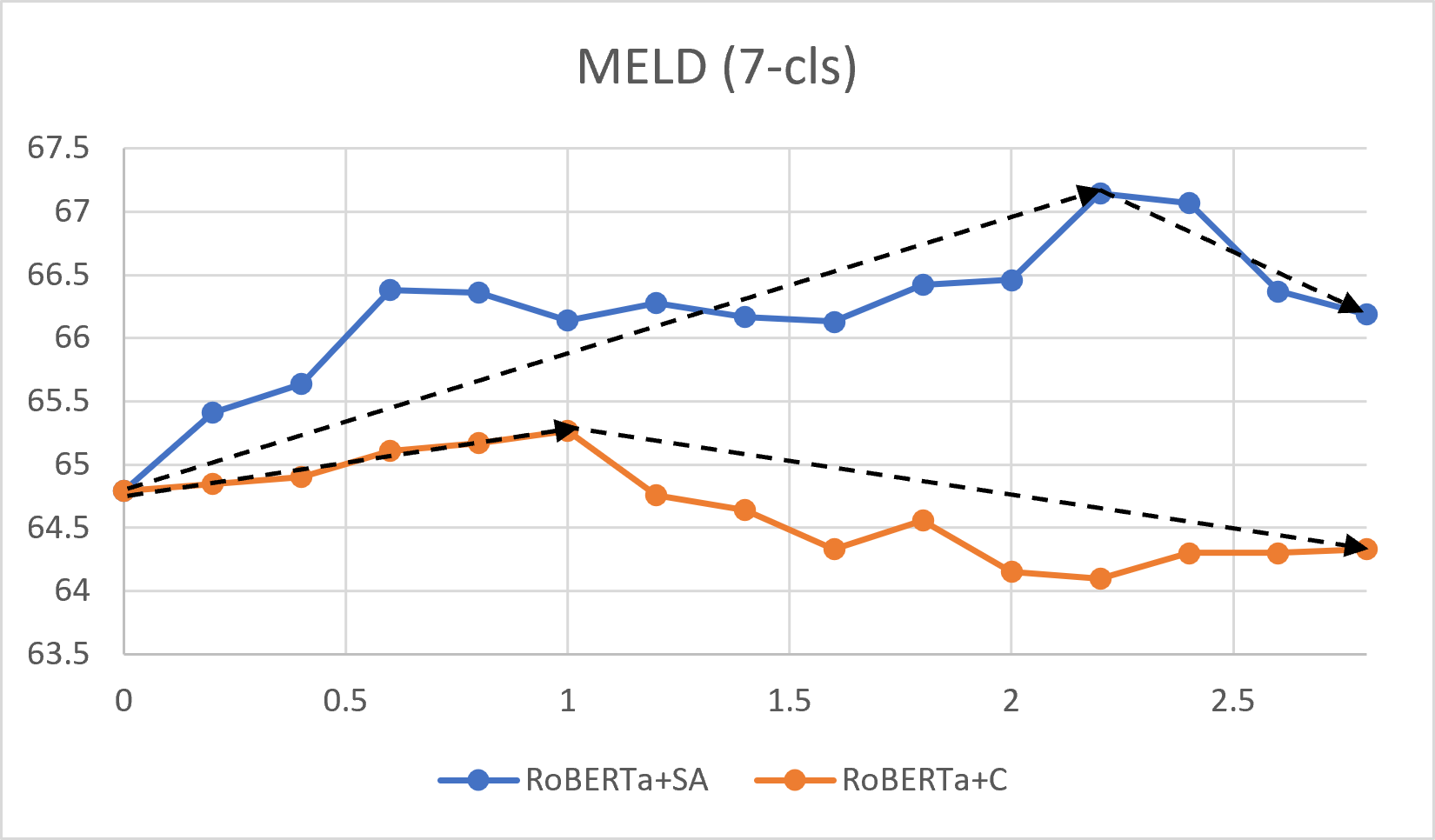}
    \caption{Performance of RoBERTa+SA and RoBERTa+C according to $\alpha$ in MELD.}
    \label{fig:alpha}
\end{figure}


This section demonstrates that the optimal $\alpha$ is determined by a combination of model structure, data, and grayscale type. When RoBERTAa is trained with MELD, the performance according to $\alpha$ for the most effective +FSA and least effective +C is measured and the results are shown in Figure~\ref{fig:alpha}. Since +FSA and +C have the largest difference in effect, it is considered intuitive to compare these two methods.

As $\alpha$ increases, RoBERTa is trained with more emphasis on the distribution of grayscale labels. $\alpha$ up to the threshold improves the performance, but when it becomes larger than the threshold, the performance decreases. The threshold depends on the type of data and grayscale. We focus on the effect of grayscale labels, not to find the optimal $\alpha$ for each combination. Therefore, $\alpha$ is used as a fixed value (=1) in our experiments. We confirm that $\alpha$ shows similar effects if it is not too small, which is verified through multiple runs. Our future work is to find the optimal $\alpha$ according to the learning environment.

\section{Conclusion}
We introduce a novel approach in ERC, which automatically constructs different types of grayscale labels taking into account correlations between emotions. The "category" and "word-embedding" methods construct a grayscale label based on the ground-truth emotion, while the other methods construct a grayscale label based on the dialogue. However, the self-model reflecting the dialogue may have incorrect information with ground-truth because it leverages the pre-trained model to generate grayscale labels. Therefore we introduce enhanced methods such as SA and FSA. We show competitive performance by learning RoBERTa as a grayscale label. It can also improve the performance of state-of-the-art models.

The different types of grayscale labels we proposed have different effects depending on the data and model structure. Also, finding the optimal $\alpha$ in the loss Equation~\ref{eq:loss} should be determined through several experiments. We will explore further in this related future study.

\bibliographystyle{IEEEtran}
\bibliography{mybib}

\appendix

\end{document}